%% file: egpaper_final.tex
\documentclass[10pt,twocolumn,letterpaper]{article}

\usepackage{iccv}
\usepackage{times}
\usepackage{epsfig}
\usepackage{graphicx}
\usepackage{amsmath}
\usepackage{amssymb}
\usepackage{subfigure}

\input{math_commands.tex}

\usepackage{times}
\usepackage{epsfig}
\usepackage{graphicx}
\usepackage{amsmath}
\usepackage{amssymb}
\usepackage{booktabs}
\usepackage{multirow}
\usepackage{url}
\usepackage{multirow}
\usepackage[ruled,linesnumbered]{algorithm2e}

\usepackage[breaklinks=true,bookmarks=false]{hyperref}

\iccvfinalcopy % *** Uncomment this line for the final submission

\def\iccvPaperID{1271} % *** Enter the ICCV Paper ID here

\ificcvfinal\fi

\begin{document}

%%%%%%%%% TITLE+
\title{Seed the Views: Hierarchical Semantic Alignment for Contrastive Representation Learning}

% % Additioneither the email address or home page, not both
% \andal authors and addresses can be added with ``\and'',
% % To save space, use
% \author{
% Haohang Xu\textsuperscript{1,2}\quad  Xiaopeng Zhang\textsuperscript{2}\quad Hao Li\textsuperscript{1,2}\quad Lingxi Xie\textsuperscript{2}\quad Hongkai Xiong \textsuperscript{1}\quad Qi Tian\textsuperscript{2}\\
% \textsuperscript{1}Shanghai Jiao Tong University,\quad\textsuperscript{2}Huawei Inc.\\
% % \small\texttt{xinyueh@mail.ustc.edu.cn}\quad\small\texttt{\{198808xc,zxphistory,wlh2568@gmail.com}\\
% % \small\texttt{lihao0374@sjtu.edu.cn}\quad\small\texttt{yangzijie@ict.ac.cn}\quad\small\texttt{\{zhwg,lihq\}@ustc.edu.cn}\quad\small\texttt{tian.qi1@huawei.com}
% \small\texttt{\{xuhaohang,lihao0374,xionghongkai\}@sjtu.edu.cn}\\\small\texttt{\{zxphistory,198808xc\}@gmail.com}\quad\small\texttt{tian.qi1@huawei.com}
% }
\author{
Haohang Xu\textsuperscript{1,2}\quad  Xiaopeng Zhang\textsuperscript{2}\quad Hao Li\textsuperscript{1,2}\quad Lingxi Xie\textsuperscript{2}\quad Hongkai Xiong \textsuperscript{1}\quad Qi Tian\textsuperscript{2}\\
\textsuperscript{1}Shanghai Jiao Tong University,\quad\textsuperscript{2}Huawei Inc.\\
% \small\texttt{xinyueh@mail.ustc.edu.cn}\quad\small\texttt{\{198808xc,zxphistory,wlh2568@gmail.com}\\
% \small\texttt{lihao0374@sjtu.edu.cn}\quad\small\texttt{yangzijie@ict.ac.cn}\quad\small\texttt{\{zhwg,lihq\}@ustc.edu.cn}\quad\small\texttt{tian.qi1@huawei.com}
\small\texttt{\{xuhaohang,lihao0374,xionghongkai\}@sjtu.edu.cn}\\\small\texttt{\{zxphistory,198808xc\}@gmail.com}\quad\small\texttt{tian.qi1@huawei.com}
}

\maketitle

%%%%%%%%% ABSTRACT
\begin{abstract}
   Self-supervised learning based on instance discrimination has shown remarkable progress. In particular, contrastive learning, which regards each image as well as its augmentations as an individual class and tries to distinguish them from all other images, has been verified effective for representation learning. However, pushing away two images that are de facto similar is suboptimal for general representation. In this paper, we propose a hierarchical semantic alignment strategy via expanding the views generated by a single image to \textbf{Cross-samples and Multi-level} representation, and models the invariance to semantically similar images in a hierarchical way. This is achieved by extending the contrastive loss to allow for multiple positives per anchor, and explicitly pulling semantically similar images/patches together at different layers of the network. Our method, termed as CsMl, has the ability to integrate multi-level visual representations across samples in a robust way. CsMl is applicable to current contrastive learning based methods and consistently improves the performance. Notably, using the moco as an instantiation, CsMl achieves a \textbf{76.6\% }top-1 accuracy with linear evaluation using ResNet-50 as backbone, and \textbf{66.7\%} and \textbf{75.1\%} top-1 accuracy with only 1\% and 10\% labels, respectively. \textbf{All these numbers set the new state-of-the-art.}
\end{abstract}

%%%%%%%%% BODY TEXT
\section{Introduction}
As a fundamental task in machine learning, representation learning targets at extracting compact features from the raw data, and has been dominated by the fully supervised paradigm over the past decades \cite{he2016deep}, \cite{russakovsky2015imagenet}, \cite{yun2019cutmix}, \cite{zhang2017mixup}. Recent progress on representation learning has witnessed a remarkable success over self-supervised learning \cite{doersch2015unsupervised}, \cite{gidaris2018unsupervised}, \cite{noroozi2016unsupervised}, \cite{pathak2016context}, \cite{wu2018unsupervised}, which facilitates feature learning without human annotated labels. In self-supervised learning, a network is trained based on a series of predefined tasks according to the intrinsic distribution priors of images, such as image colorization \cite{zhang2016colorful}, rotation prediction \cite{gidaris2018unsupervised}, context completion \cite{pathak2016context}, \emph{etc}. More recently, the contrastive learning method \cite{he2020momentum}, which is based on instance discrimination as a pretext task, has taken-off as it has been demonstrated to outperform the supervised counterparts on several downstream tasks like classification and detection.

\begin{figure}[t!]
  \begin{center}
        \includegraphics[width=0.98\linewidth,height=5cm]{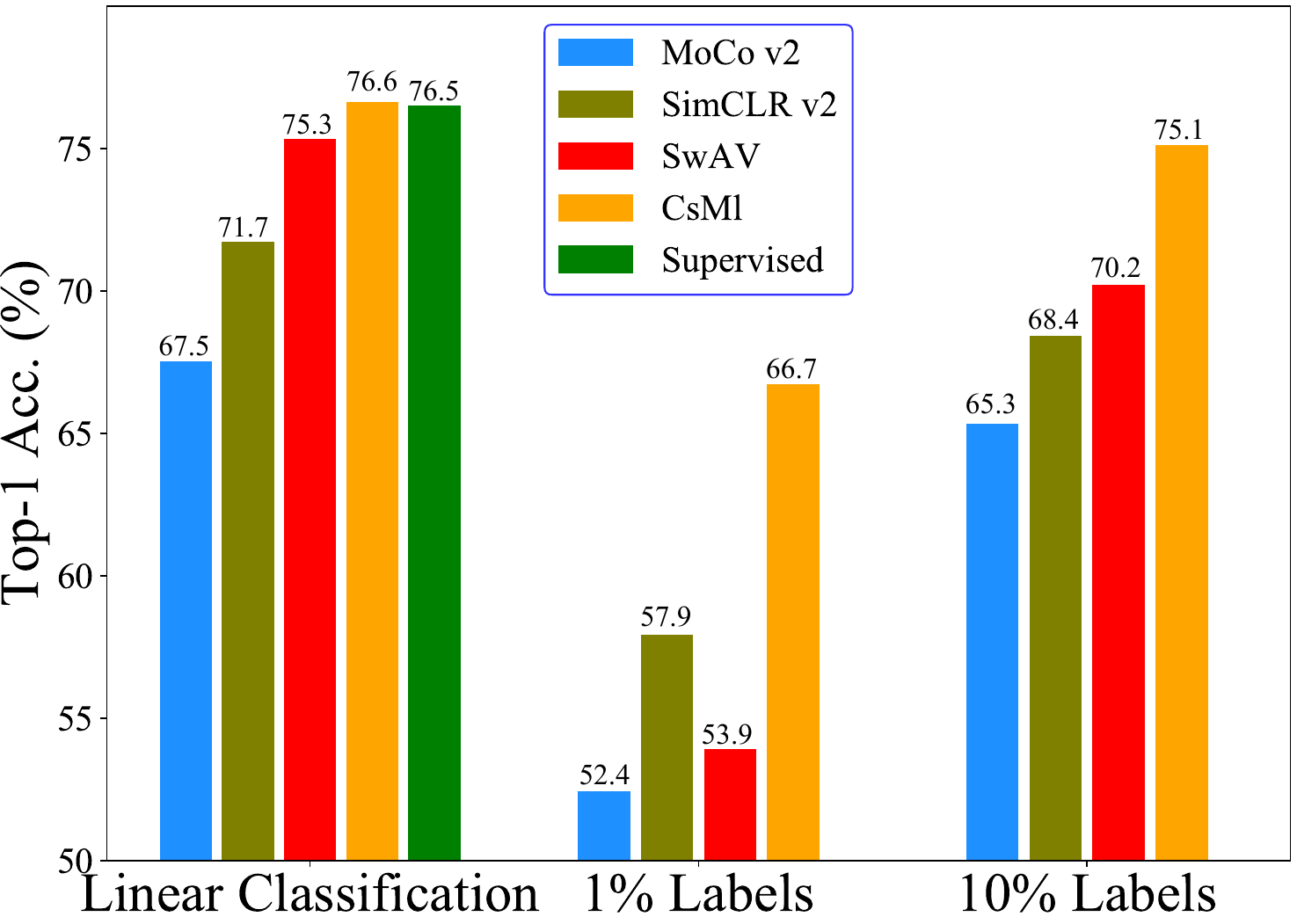}
  \end{center}
  \vspace{-0.2cm}
     \caption{An overall performance comparison of our proposed pre-trained features on several widely evaluated unsupervised benchmarks. Here, a standard ResNet-50 is used as backbone. CsMl significantly improves the performance on various tasks.}
  \label{overall_results}
\end{figure}

In contrastive learning, each image as well its augmentations is treated as a separate class, and the views generated by a single image are pulled closer together, while all other images are treated as negatives and pushed away. In this setting, the invariance is only encoded from low-level image transformations such as cropping, blurring, and color jittering, \emph{etc}. While the invariance to semantically similar images is not explicitly modeled but on the contrary, they are treated as negatives and pushed away as in MoCo \cite{he2020momentum}. This is contradictory with the alignment principle \cite{wang2020understanding} of feature representation, which favors the encoders to assign similar features to similar samples. As a result, the optimization is contradictory with the intrinsic distribution of images and is not optimal for feature representation. %This is especially true for the intermediate layers that accept gradients from the last embedding layers.

%Motivated by the scattered feature distribution of contrastive learning, this paper targets at pursuing a more compact representation under unsupervised setting.
In this paper, we introduce a hierarchical semantic alignment strategy via seed the views that are constrained within a single image and a single level representation to \textbf{Cross-samples and Multi-levels}. The idea behind our strategy is to align semantically similar samples in different latent space, and thus enable better representation throughout the network. Specifically, the cross-sample views are achieved by simply searching the feature representation in the embedding space, and selecting the nearest neighbors that are similar with the anchor for contrastive learning. While the multi-level views are expanded at the intermediate layers of a network, which enables hierarchical representation of the same image/patch. As a result, the views are seeded from a single image, and expanded across different samples at different levels, these views are pulled together for more general and discriminative representation.

For \textbf{cross-sample views}, it is a dilemma to select appropriate nearest samples as positives, pulling samples that are \emph{de facto} very similar in the feature space brings about limited performance gain since current representation handles these invariance well, while enlarging the searching space would inevitably introduce noisy samples. To solve this issue, we rely on data mixing to generate extra positive samples, which can be treated as a smoothing regularization of the anchor. In this way, similar samples are extended in a smoother and robust way, and we are able to better model the intra-class similarity for compact representation. %Furthermore,guided by the alignment principle, we extend the semantic alignment to earlier layers and propose a hierarchical training strategy that enables the feature representation to be compact throughout the network. This is achieved by explicitly enforcing contrastive loss at intermediate representation layers as well as the last embedding layers. We find that better semantic alignment in the middle layers also benefits for transferability.

%due to the representation gap between feature space and semantic space, two images that are similar in feature representation may not correspond to the same labels in semantic space, and inevitably introducing noisy positive samples. To solve this issue, we introa data mixing as extra positive samples, which can be treated as a smoothing version of the anchor. In particular, we apply Cutmix \cite{yun2019cutmix}, a widely used data augmentation in supervised learning, where patches are cut and pasted among the positive pairs to generate new samples. Benefit from the center-wise sample selection, the Cutmix augmentation is only constrained within the local neighborhood of an image, and can be treated as an expansion of current neighborhood space. In this way, similar samples are pulled together in a smoother and robust way, which we find is beneficial for general representation.

%This is motivated by the design principle of feature representation, \emph{i.e.,} a good representation should satisfy both alignment and uniformity []. Alignment favors encoders that assign similar features to similar samples, and uniformity prefers feature vectors roughly uniformly distributed on the unit hypersphere, and is beneficial from well-clustered compactness distributions.
%Previous works \cite{khosla2020supervised} [debiased contrastive learning] tackle the via with the
For \textbf{multi-level views}, we find that although the linear classification accuracy of the last layers is approaching the supervised baseline \cite{chen2020simple}, \cite{chen2020big}, the middle-level representation of current contrastive based methods suffers much lower discrimination capacity, which is harmful for downstream tasks such as detection that require intermediate discrimination ability. Towards this goal, we extend the views to intermediate layers of a network and propose a hierarchical training strategy that enables the feature representation to be more discriminative in the intermediate layers. However, it suffers from optimization contradiction when directly adding a loss layer on the intermediate representation due to the gradient competition issue. To solve this issue, we add a bottleneck layer for each intermediate loss, which we find is applicable for robust optimization. In this way, the features are deeply supervised throughout the network, which also benefits for transferability.

%due to the representation gap between feature space and semantic space, two images that are similar in feature representation may not correspond to the same labels in semantic space, and inevitably introducing noisy positive samples. To solve this issue, we introa data mixing as extra positive samples, which can be treated as a smoothing version of the anchor. In particular, we apply Cutmix \cite{yun2019cutmix}, a widely used data augmentation in supervised learning, where patches are cut and pasted among the positive pairs to generate new samples. Benefit from the center-wise sample selection, the Cutmix augmentation is only constrained within the local neighborhood of an image, and can be treated as an expansion of current neighborhood space. In this way, similar samples are pulled together in a smoother and robust way, which we find is beneficial for general representation.

%This is motivated by the design principle of feature representation, \emph{i.e.,} a good representation should satisfy both alignment and uniformity []. Alignment favors encoders that assign similar features to similar samples, and uniformity prefers feature vectors roughly uniformly distributed on the unit hypersphere, and is beneficial from well-clustered compactness distributions.
%Previous works \cite{khosla2020supervised} [debiased contrastive learning] tackle the via with the

The proposed hierarchical semantic alignment strategy via CsMl significantly boosts the feature representation of contrastive learning when evaluated on several self-supervised learning benchmarks. As shown in Fig. \ref{overall_results}, using MoCo as an instantiation  \cite{he2020momentum}, we achieve $76.6\%$ top-1 accuracy with a standard ResNet-50 on ImageNet linear evaluation, and $66.7\%$ and $75.1\%$ top-1 accuracy with $1\%$ and $10\%$ labels, respectively. We also validate its transferring ability on several downstream tasks covering detection and segmentation, and achieve better results comparing with previous self-supervised methods.

\section{Related Work}
Self-supervised representation learning has attracted more and more attentions over the past few years since it is free of labels and is easy to scale up. Self-supervised learning aims at exploring the intrinsic distribution of data samples via constructing a series of pretext tasks, which varies in utilizing different priors of images. Traditional self-supervised learning has sought to learn a compressed code which can effectively reconstruct the input. Among them, a typical strategy is to take advantage of the spatial properties of images, like predicting the relative spatial positions of images patches \cite{doersch2015unsupervised}, \cite{noroozi2016unsupervised}, or inferring the missing parts of images by inpainting \cite{pathak2016context}, colorization \cite{zhang2016colorful}, or rotation prediction \cite{gidaris2018unsupervised} \emph{etc.}. Recent progress in self-supervised learning is mainly based on instance discrimination \cite{wu2018unsupervised}, in which each image as well as its augmentations is treated as a separate class. The motivation behind these works is the InfoMax principle, which aims at maximizing mutual information \cite{tian2019contrastive}, \cite{wu2018unsupervised} across different augmentations of the same image \cite{chen2020simple}, \cite{he2020momentum}, \cite{tian2019contrastive}. The design choices of the InfoMax principle, such as the number of negatives and how to sample them, hyper-parameter settings, and data augmentations all play a critical role for a good representation.

%The key idea is that different augmentations or different views of an image should map to similar embedding that is unique across image samples.

Data augmentation plays a key role for contrastive learning based representation. Since the invariance is only encoded by different transformations of an image. According to \cite{chen2020simple}, \cite{tian2020makes}, the performance of contrastive learning based approaches strongly relies on the types and strength of augmentations, \emph{i.e.}, image transformation priors that do not change object identity. In this way, the network is encouraged to hold invariance in the local vicinities of each sample, and usually more augmentations benefit for feature representation. However, current widely used data augmentation methods are mostly operated within a single sample. One exception is the method in \cite{shen2020rethinking}, which makes use of mixup mixture for flattened contrastive predictions. However, such a mixture strategy is conducted among all the images, which destroys the local similarity when contrasting mixed samples that are semantically dissimilar.

Our method is reminiscent of the recent proposed Supervised Contrastive Learning \cite{khosla2020supervised}, which pulls multiple positive samples together. The differences are that, first, SCL is designed for fully supervised paradigm, where the positive samples are simply selected from the ground truth labels, while our method does not rely on these labels, and deliberately design a positive sample selection strategy to facilitate semantic alignment in a robust way. Second, we extend the contrastive loss to the intermediate hidden layers to enhance its discriminative power of the earlier layers, which is beneficial for discriminative representation and with better transferability, especially for semi-supervised learning.

\begin{figure*}[t!]
  \begin{center}
        \includegraphics[width=0.9\linewidth,height=7.6cm]{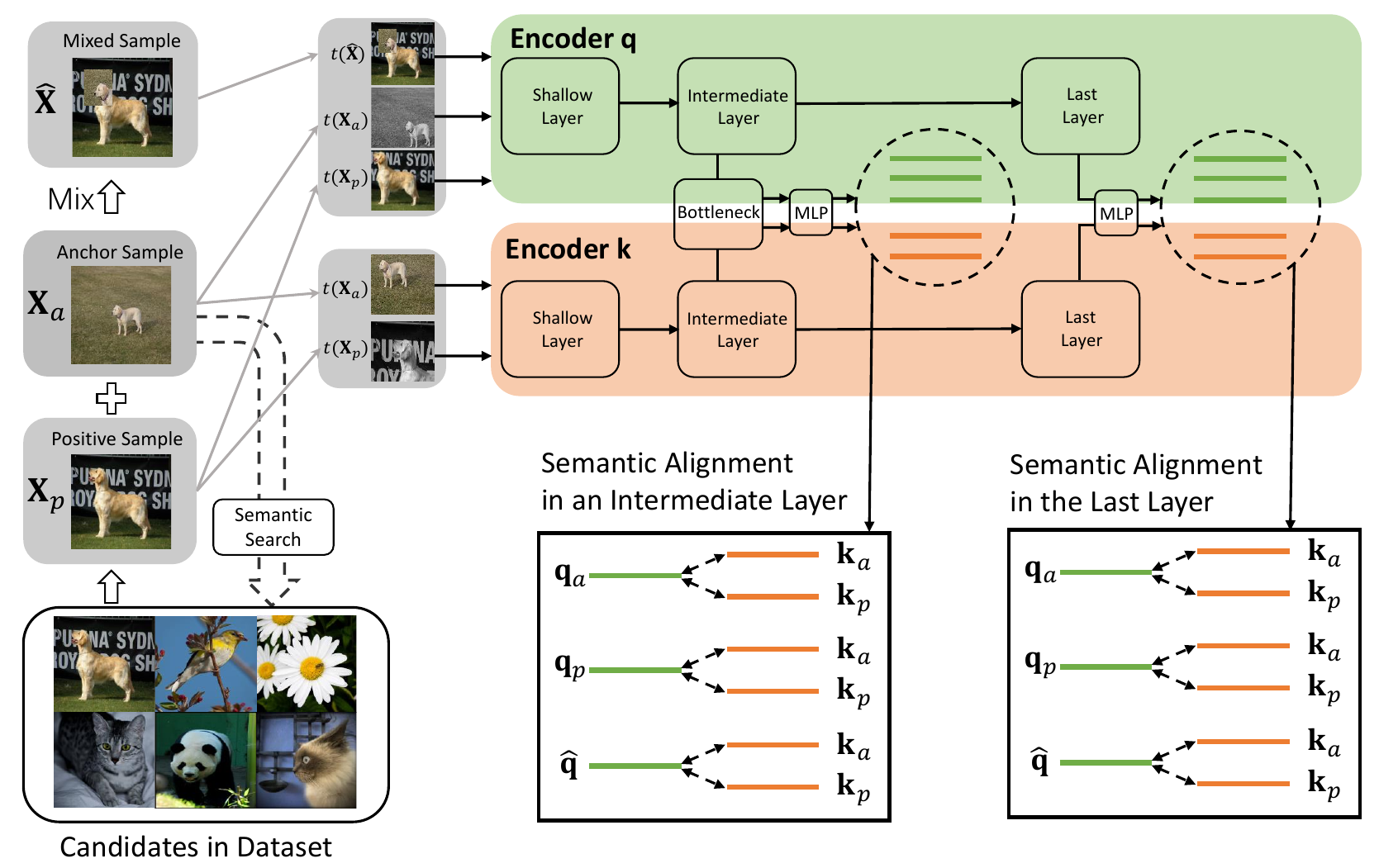}
  \end{center}
  \vspace{-0.3cm}
     \caption{An overview of our proposed hierarchical semantic alignment framework. The baseline is based on MoCo \cite{he2020momentum}, which requires a query encoder $f_{\rm q}$ and an asynchronously updated key encoder $f_{\rm k}$. Given an anchor image $\mathbf{x}_a$, we randomly select a positive sample $\mathbf{x}_p$ from the nearest neighborhood set $\Omega$, and generate the mixed sample $\mathbf{\hat{x}}$. The hierarchical semantic alignment is enforced by pulling the positive samples $\mathbf{x}_a$, $\mathbf{x}_p$, and $\mathbf{\hat{x}}$ in the intermediate layers as well as the last embedding space.}
  \label{framework}
\end{figure*}

\section{Methodology}
In this section, we start by reviewing contrastive loss for self-supervised learning, and investigate its drawbacks for general feature representation. Then we present the proposed hierarchical training strategy that pulls semantically similar images at different layers of a network. As shown in Fig. \ref{framework}, the core idea includes two modules, first, we deliberately design a positive sample selection strategy to expand the neighborhood of a single image, and adjust the contrastive loss to allow for multiple positives during each forward propagation. Furthermore, we propagate the semantic alignment to the earlier layers to encourage class separability. In this way, the features are trained in a hierarchical way for more compact representation. Each module would be elaborated in the following.

\subsection{An Overview of Contrastive Learning}

Contrastive learning targets at learning an encoder that is able to map positive pairs to similar representations while pushing away those negative samples in the embedding space. It can be efficiently addressed via momentum contrast \cite{he2020momentum}, which substantially increases the number of negative samples. Given a reference image with two augmented views $\mathbf{x}$ and $\mathbf{x'}$, MoCo aims to learn feature representation $\mathbf{q}=f_{\rm q}(\mathbf{x})$ by a query encoder $f_{\rm q}$, that can distinguish $\mathbf{x'}$ from all other images $\mathbf{x_i}$, where $\mathbf{x'}$ and all the negatives $\mathbf{x_i}$ are encoded by an asynchronously updated key encoder $f_{\rm k}$, with $\mathbf{k^+}=f_{\rm k}(\mathbf{x'})$ and $\mathbf{k_i}=f_{\rm k}(\mathbf{x_i})$, the contrastive loss can be defined as:

\begin{equation}\label{moco}
{\mathcal{L}_{\mathbf{q}}} \!=\!  - \log \frac{{\exp (\mathbf{q} \cdot {\mathbf{k^+} }/\tau )}}{{\sum_{i=0}^{N} \exp (\mathbf{q} \cdot {\mathbf{k_i} }/\tau )}},
\end{equation}

\noindent where $\tau$ is the temperature parameter scaling the distribution of distances. However, the positive samples are constrained within a single image with different transformations, and only support one positive sample for each query $\mathbf{q}$, which is low efficient and hard for modeling invariance to semantically similar images.

\subsection{Contrastive Learning with Cross-Sample Views}
In this section, we extend the views generated from a single image to cross-sample views, and describe the proposed positive sample selection strategy and adjust contrastive loss to allow for multiple positives to explicitly model the invariance among similar images.

\textbf{Positive Sample Selection.} For positive samples, we simply make use of $k$ nearest neighbors to search semantically similar images in the embedding space. Specially, given unlabeled training set $\bm{X}=\left \{\mathbf{x_{1}},\mathbf{x_{2}},...,\mathbf{x_{n}}\right \}$ and a query encoder $f_{\rm q}$, we obtain the corresponding embedding representation $\bm{V}=\left \{ \mathbf{v_{1}},\mathbf{v_{2}},...,\mathbf{v_{n}} \right \}$ where $\mathbf{v_i} = f_{\rm q}(\mathbf{x_i})$. For a typical Res-50 network, the embedding is obtained from the last average pooled features with dimension 2048. Given an anchor sample $\mathbf{x}_a$, we compute the cosine similarity with all other images, and  select the top $k$ samples with the highest similarity as positives $\Omega = \{\mathbf{x^1}, \mathbf{x^2}, ..., \mathbf{x^k}\}$.

\textbf{Loss Function.} We simply adjust the contrastive loss in Eq. \ref{moco} to allow for multiple positives per anchor. Given an anchor sample $\mathbf{x}_a$ and its nearest neighborhood set $\Omega$, we randomly select a positive sample $\mathbf{x}_p \in \Omega$, and the loss term $\mathcal{L}_{\mathbf{q}_a}$ for $\mathbf{x}_a$ can be reformulated as:

\begin{align}\label{eq: q_a_loss_term}
\mathcal{L}_{\mathbf{q}_a} \!=\!  - \frac{1}{2} \left[\log \frac{\exp(\mathbf{q}_{a} \cdot \mathbf{k}_{a}/\tau)}{\exp(\mathbf{q}_{a} \cdot \mathbf{k}_{a} /\tau) + \sum_{i=1}^N \exp(\mathbf{q}_a\cdot \mathbf{k_i} /\tau)}\right. \notag \\
+ \left.\log \frac{\exp(\mathbf{q}_a \cdot \mathbf{k}_p/\tau)}{\exp(\mathbf{q}_a \cdot \mathbf{k}_p/\tau) + \sum_{i=1}^N \exp(\mathbf{q}_a \cdot \mathbf{k_i} /\tau)}\right],
\end{align}
where each anchor sample $\mathbf{q}_a$ encoded with $f_{\rm q}$, is pulled with two samples $\mathbf{k}_a$ and $\mathbf{k}_q$ encoded with $f_{\rm k}$, and pushed away with all other samples in the key encoder $f_{\rm k}$. Symmetrically, the loss term $\mathcal{L}_{\mathbf{q}_p}$ for positive sample $\mathbf{x}_p$ can be obtained accordingly. The overall loss is the combination of the two losses, which is equipped with two positive samples in the query encoder $f_{\rm q}$, and the corresponding two positive samples in the key encoder $f_{\rm k}$. Each sample is accompanied with a random data augmentation as described in \cite{chen2020improved}, and is pulled together with all positive samples (also undergo a random data augmentation) from the other encoder.

%\begin{align}\label{eq: q_p_loss_term}
%\mathcal{L}_{q_p} \!=\! -\frac{1}{2} (\log \frac{\exp(q_p %\cdot k_p/\tau)}{\exp(q_p \cdot k_p /\tau) + \sum_{i=1}^K %\exp(q_p \cdot k_i /\tau)} \notag \\
%+ \log \frac{\exp{(q_p \cdot k_a/\tau)}}{\exp(q_p \cdot k_a %/\tau) + \sum_{i=1}^K \exp{(q_p \cdot k_i/\tau)}}).
%\end{align}

\textbf{Expanding the Neighborhood.} It is a dilemma to define an appropriate $k$ for nearest sample selection, setting it too small, the objective pulls samples that are already very similar in the feature space, and brings about limited performance gain since current representation handles these invariance well, while setting it too large, it would inevitably introduce noisy samples, and pulling these samples would destroy the local similarity constraint and is harmful for general representation. To solve this issue, we rely on data mixture to expand the neighborhood space of an anchor based on the selected positive samples. The assumption is that the mixed samples act as an interpolation between two samples, and lies in the local neighborhood of the two samples. In this way, the generated mixed samples expand the neighbors in the embedding space that current model cannot handle well, and pulling these samples is beneficial for better generalization.

%The motivation is that we hope to pull samples that are similar in the semantic space but are relatively far away in the feature space for better generalization.
In particular, we apply CutMix \cite{yun2019cutmix} augmentation, which is widely used as a regularization strategy to train neural networks in fully supervised paradigm. Given an anchor sample $\mathbf{x}_a$, and its positive neighbor $\mathbf{x}_{p} \in \Omega$, the mixed sample $\mathbf{\hat{x}}$ is generated as follows:
\begin{align}
    \mathbf{\hat{x}} = \mathbf{M} \odot \mathbf{x}_{a} + (\mathbf{1}-\mathbf{M}) \odot \mathbf{x}_{p},
\end{align}
where $\mathbf{M} \in \{0,1\}^{W\times H}$ is a binary mask that has the same size as $\mathbf{x}_a$, and indicates where to drop out the region in $\mathbf{x}_a$ and replaced with a randomly selected patch from $\mathbf{x}_p$, and $W,H$ denotes the width and height of an image, respectively. $\mathbf{1}$ is a binary mask filled with ones, and $\odot$ is the element-wise multiplication operation. For mask $\mathbf{M}$ generation, we simply follow the setting in \cite{yun2019cutmix}, and do not carefully tune the parameters. Note that different from CutMix used in fully supervised learning that randomly selects two images for mixing and changes the corresponding labels accordingly, we only sample those similar samples and ensure that the generated samples lie in the local neighborhood of the anchor.
% We claim that \emph{neighborhood mixture is the key for success for self-supervised learning} and would validate its effectiveness in the experimental section.

The mixed samples $\mathbf{\hat{x}}$ is treated as a new positive sample, and pulled together with $\mathbf{x}_a$ and $\mathbf{x}_p$ accordingly:

\begin{align}\label{eq: mix_loss_term}
    \mathcal{L}_{\mathbf{\hat{q}}} \!=\! -\left[\lambda \log \frac{\exp(\mathbf{\hat{q}}\cdot \mathbf{k}_a/\tau)}{\exp(\mathbf{\hat{q}}\cdot \mathbf{k}_a/\tau) + \sum_{i=1}^N \exp(\mathbf{\hat{q}}\cdot \mathbf{k_i}/\tau)}\right. \notag \\
    + \left.(1-\lambda)\log \frac{\exp(\mathbf{\hat{q}}\cdot \mathbf{k_p}/\tau)}{\exp(\mathbf{\hat{q}}\cdot \mathbf{k_p} /\tau) + \sum_{i=1}^N \exp(\mathbf{\hat{q}}\cdot \mathbf{k_i}/\tau)}\right],
\end{align}
where $\lambda$ is a combination ratio that determines the cropped area for CutMix operation, and is sampled from beta distribution Beta$(\alpha,\alpha)$ with parameter $\alpha$ ($\alpha=1$). The final loss function can be formulated as :

\begin{align}\label{final_loss}
    \mathcal{L} = \mathcal{L}_{\mathbf{q}_a} + \mathcal{L}_{\mathbf{q}_p} + \mathcal{L}_{\mathbf{\hat{q}}}.
\end{align}

We simply set the balancing factors of the three terms as 1 in all our experiments.

\subsection{Contrastive Learning with Multi-level Views}
Following training a customized network, the contrastive loss is only penalized at the last embedding layers, while the optimization of the intermediate hidden layers is implicitly penalized by back propagating the gradients to the earlier layers. However, due to the lack of labels, the optimization objective is more challenging and suffers slow convergence, and the intermediate layers are especially under fitted and with limited discriminative power comparing with fully supervised learning (c.f. Fig. \ref{fig:inter_layer}). Inspired by \cite{lee2015deeply}, we extend the proposed contrastive loss in Eq. \ref{final_loss} to the intermediate hidden layers, which targets at explicitly modeling the similarities among image/patches for better discrimination. Specifically, we introduce a companied objective at the end of each stage for a typical ResNet network, which acts as an additional constraint during the optimization procedure.

As demonstrated in \cite{romero2014fitnets}, it is too aggressive to directly add a loss layer as side branch of the intermediate layers due to the gradient competition issue. The optimization objective from one side branch would probably inconsistent with that from other branches since they undergoes extremely different levels of layers during back propagation. To solve this issue, for each companied loss at stage $l$, we add another embedding layer $g^l$ which consists of bottleneck layers before the contrastive loss as side branch. In practice, we find that a single bottleneck layer is applicable to alleviate the gradient competition and enable efficient optimization. \emph{Note that These companied branches are removed after training and hence do not increase complexity of the network after pretraining}.

% bottleneck layer $g^l$ before the contrastive loss.
Specifically, given encoder $f_{\rm q}$ and $f_{\rm k}$, we define feature map at stage $l$ as $f_{\rm q}^l$ and $f_{\rm k}^l$, which is truncated at stage $l$. Each encoder is passed through another embedding layer $g^l$, which consists of a bottleneck layer and 2 MLP layers to encode the feature into a 128-dim vector. The loss function of $q_a$ specific to stage $l$ is defined as:

\begin{align}\label{eq:mulhead_qa_loss_term}
\mathcal{L}_{\mathbf{q}_a}^l \!=\!  - \frac{1}{2} \left[\log \frac{\exp(\mathbf{q}_a^l \cdot \mathbf{k}_a^l/\tau)}{\exp(\mathbf{q}_a^l \cdot \mathbf{k}_a^l /\tau) + \sum_{i=1}^N \exp(\mathbf{q}_a^l\cdot \mathbf{k_i}^l /\tau)}\right. \notag \\
+ \left.\log \frac{\exp(\mathbf{q}_a^l \cdot \mathbf{k}_p^l/\tau)}{\exp(\mathbf{q}_a^l \cdot \mathbf{k}_p^l/\tau) + \sum_{i=1}^N \exp(\mathbf{q}_a^l \cdot \mathbf{k_i}^l /\tau)}\right],
\end{align}
where $\mathbf{q}^l = g^l(f_{\rm q}^l)$ and $\mathbf{k}^l = g^l(f_{\rm k}^l)$. Similarly,  the loss term $\mathcal{L}^l_{\mathbf{\hat{q}}}$ and $\mathcal{L}^l$ can be obtained according to Eq. \ref{eq: mix_loss_term} and Eq. \ref{final_loss}, respectively. When there are $L$ losses corresponding to $L$ intermediate stages, the final losses of the whole network can be computed as:
\begin{align}
    \mathcal{L}_{total} = \mathcal{L}+\sum_{l=1}^L \mathcal{L}^l
\end{align}

\begin{table}[]
\caption{Top-1 accuracy under linear classification on ImageNet with ResNet-50 as backbone.}
\vspace{0.05in}
\fontsize{10}{12}\selectfont
\centering
\setlength{\tabcolsep}{3mm}
\begin{tabular}{lc}
\toprule
Method     & Accuracy(\%) \\
%\midrule
%Supervised & 76.5   \\
\midrule
%Colorization\cite{zhang2016colorful}    & 39.6    \\
%Jigsaw Puzzles\cite{noroozi2016unsupervised}  & 45.7      \\
%NPID \cite{wu2018unsupervised} &54.0 \\
BigBiGAN\cite{donahue2019large} &56.6 \\
Local aggregation \cite{zhuang2019local} &58.8 \\
%Moco \cite{he2020momentum} &60.6 \\
SeLa \cite{asano2020self} &61.5 \\
PIRL \cite{misra2020self} &63.6 \\
CPCv2 \cite{henaff2019data} &63.8 \\
PCL \cite{li2020prototypical} &65.9 \\
%SimCLR \cite{chen2020simple} &70.0 \\
SimCLRv2 \cite{chen2020big} &71.7 \\
MoCo v2 \cite{chen2020improved} &71.1 \\
BYOL \cite{grill2020bootstrap} &74.3 \\
SwAV \cite{caron2020unsupervised} &75.3 \\
\midrule
CsMl 200 Epochs (w/o multi-crop) & 71.6 \\
CsMl 800 Epochs (w/o multi-crop) &74.4 \\
\midrule
CsMl 200 Epochs(w/ multi-crop) & 74.6 \\
% \textcolor{red}{CsMl-BYOL} & \textcolor{red}{75.3} \\
CsMl 800 Epochs(w/ multi-crop) & \textbf{76.6} \\
%CsMl+Multi-crop & \textbf{76.6} \\
%CsMl(1200 Epoch) & \textbf{76.5} \\
\bottomrule
\end{tabular}
\label{tab:lincls_imagenet1K}
\end{table}

% \begin{table}[]
% \caption{Top-1 accuracy under linear classification on ImageNet with ResNet-50 as backbone.}
% \vspace{0.02in}
% % \fontsize{10}{12}\selectfont
% \centering
% \setlength{\tabcolsep}{4mm}
% \begin{tabular}{lcc}
% \toprule
% Method     & 200 epochs  & 800 epochs\\
% %\midrule
% %Supervised & 76.5   \\
% \midrule
% %Colorization\cite{zhang2016colorful}    & 39.6    \\
% %Jigsaw Puzzles\cite{noroozi2016unsupervised}  & 45.7      \\
% %NPID \cite{wu2018unsupervised} &54.0 \\
% % BigBiGAN\cite{donahue2019large} &56.6 \\
% % Local aggregation 200epochs \cite{zhuang2019local} &58.8 \\
% %Moco \cite{he2020momentum} &60.6 \\
% % SeLa \cite{asano2020self} &61.5 \\
% PIRL \cite{misra2020self} &63.6 \\
% CPCv2 \cite{henaff2019data} &63.8 \\
% PCL \cite{li2020prototypical} &65.9 \\
% SimCLR \cite{chen2020simple} &66.5 &70.0 \\
% SimCLRv2 \cite{chen2020big} &71.7 \\
% MoCo v2  \cite{chen2020improved} &67.5   &71.1\\
% BYOL \cite{grill2020bootstrap} & - &74.3\\
% SwAV \cite{caron2020unsupervised} &73.9 &75.3 \\
% \midrule
% CsMl(w/ multi-crop) & 74.6  &76.6\\
% % \textcolor{red}{CsMl-BYOL} & \textcolor{red}{75.3} \\
% % CsMl 800 Epochs & \textbf{76.6} \\
% %CsMl+Multi-crop & \textbf{76.6} \\
% %CsMl(1200 Epoch) & \textbf{76.5} \\
% \bottomrule
% \end{tabular}
% \label{tab:lincls_imagenet1K}
% \end{table}
%\vspace{-0.15in}
\section{Experiments}
In this section, we access our proposed feature representation on several widely used unsupervised benchmarks. We first evaluate the classification performance on ImageNet under linear evaluation and semi-supervised protocols \cite{chen2020simple}, \cite{he2020momentum}. Then we transfer the representation to several downstream tasks including detection and instance segmentation. We also analyze the performance of our feature representation with detailed ablation studies.

\begin{table}[t]
\caption{KNN classification accuracy on ImageNet. We report top-1 accuracy with 20 and 200 nearest neighbors, respectively.}
\vspace{0.05in}
\renewcommand\arraystretch{1.2}
\centering
\setlength{\tabcolsep}{5mm}
\begin{tabular}{lcc}
\toprule
Method       & 20-NN    & 200-NN               \\
\midrule
Supervised   & 75.0     & 73.2                \\
\midrule
NPID \cite{wu2018unsupervised}        & -        & 46.5                 \\
LA\cite{zhuang2019local}           & -        & 49.4                 \\
PCL\cite{li2020prototypical}          & 54.5     & -                    \\
MoCo v2\cite{chen2020improved}      & 62.0     & 59.0                 \\
SwAV\cite{caron2020unsupervised}         & 65.7     & 62.7                 \\
\midrule
CsMl          & \textbf{72.4}     & \textbf{70.7} \\
\bottomrule
\end{tabular}
\label{tab: knn_acc}
\end{table}

\subsection{Pre-training Details}
The feature representation is trained based on a standard ResNet-50 \cite{he2016deep} network, using ImageNet 2012 training dataset \cite{deng2009imagenet}. We follow the settings MoCo as in \cite{he2020momentum} \footnote{Our method is also applicable for other contrastive based method, see the appendix for the results of BYOL \cite{grill2020bootstrap}}, which employs an asynchronously updated key encoder to enlarge the capacity of negative samples, and add a 2-layer MLP on top of the last average pooling layer to form a 128-d embedding vector \cite{chen2020improved}. The model is trained using SGD optimizer with momentum 0.9 and weight decay 0.0001. The batch size and learning rate are set to 1024 and 0.12 for 32 GPUs, according to the parameters recommended by \cite{chen2020improved}, \cite{goyal2017accurate}. The learning rate is decayed to $0$ by cosine scheduler \cite{loshchilov2016sgdr} during the whole training process.

For companied loss in the intermediate stages, each branch is added with a bottleneck with three layers ($1\times 1$, $3\times 3$, and $1\times 1$ convolutions, and the number of channels follows the setting of the corresponding stage) and 2-MLP layers. We add the companied loss at both stage 2 and stage 3, and simply set the balance factor of each layer as 1.  For positive sample selection, we perform knn every 5 epochs and select top-10 nearest neighbors for each anchor. We find that the update frequency does not affect the performance too much when it ranges from 1 to 20. For efficiency, we only pull one positive sample per anchor as well as another mixed sample during each forward propagation, which we find is sufficient (see appendix for pulling more positive samples at once). For data augmentation, except for those used in \cite{he2020momentum}, we also report the results of multi-crop augmentation \cite{caron2020unsupervised}, which has been demonstrated to be effective for further performance gain. The final model is trained for 800 epochs for evaluation.

\subsection{Experiments on ImageNet}
\paragraph{Classification with Linear Evaluation.} We first evaluate our pretrained features by training a linear classifier on top of the frozen representation, following a common protocol in \cite{he2020momentum}. The classifier is trained on global average pooled features of ResNet-50 for 100 epochs, and we report the center crop, top-1 classification accuracy on ImageNet validation set. As shown in Table \ref{tab:lincls_imagenet1K}, for fair comparison, all results are based on the same network structure with the same amount of parameters. CsMl achieves $74.4\%$ top-1 accuracy under 800 epochs pretraining, which outperforms the MoCo v2 baseline \cite{chen2020improved} by $3.3\%$, and the performance can be further boosted to $76.6\%$ when adding multi-crop data augmentations, which is better than previous best performed result SwAV by $1.3\%$. %Notably, the linear classification accuracy is already on par with the fully supervised counterparts with accuracy of $76.5\%$.

\begin{table}[t]
\renewcommand\arraystretch{1.2}
\caption{Semi-supervised learning by fine-tuning $1\%$ and $10\%$ labeled images on ImageNet. The last row reports results of using a simple data mining procedure (averaged over 5 trials).}
\vspace{0.05in}
\centering
\setlength{\tabcolsep}{2.2mm}
\begin{tabular}{lcccc}
\toprule
\multicolumn{1}{c}{\multirow{2}{*}{Method}} & \multicolumn{2}{c}{$1\%$ labels} & \multicolumn{2}{c}{$10\%$ labels} \\
\multicolumn{1}{c}{}            & Top-1  & Top-5    & Top-1  & Top-5 \\\toprule
Supervised                      & 25.4   & 48.4     & 56.4   & 80.4  \\ \midrule
UDA\cite{xie2019unsupervised}   & -      & -        & 68.8   & 88.5  \\
FixMatch\cite{sohn2020fixmatch} & -      & -        & 71.5   & 89.1  \\
PIRL\cite{misra2020self}        & 30.7   & 57.2     & 60.4   & 83.8  \\
PCL\cite{li2020prototypical}    & -      & 75.6     & -      & 86.2 \\
SimCLR\cite{chen2020simple}     & 48.3   & 75.5     & 65.6   & 87.8 \\
MoCo v2\cite{chen2020improved}  & 52.4   & 78.4     & 65.3   & 86.6 \\
SwAV\cite{caron2020unsupervised}        & 53.9   & 78.5     & 70.2   & 89.9  \\
SimCLRv2\cite{chen2020big}      & 57.9   & 82.5     & 68.4   & 89.2 \\
\midrule
CsMl                            & \textbf{62.2}   & \textbf{83.0}     & \textbf{72.9}   &\textbf{90.7}   \\
CsMl+data mining                            & \textbf{66.7}   & \textbf{87.7}     & \textbf{75.1}   &\textbf{92.1}   \\
\bottomrule
\end{tabular}
\label{tab:semi_imagenet}
\end{table}

\paragraph{Classification with KNN Classifier.} We also evaluate our representation with KNN classifier, which is able to evaluate the pre-trained features more directly. Following \cite{caron2020unsupervised}, we center crop the images to obtain features from the last average pooled layers, and report the accuracy with 20 and 200 NN (we choose the result of 800 epochs with multi-crop augmentation) in Table \ref{tab: knn_acc}. For convenient comparison, we also list the KNN classification results of fully supervised model, which achieves accuracy of $75.0\%$. CsMl is only $2.6\%$ lower than the supervised baseline, and significantly outperforms previous methods, which validates the effectiveness of explicitly modeling similarities cross samples.

\paragraph{Semi-supervised Settings.} We also evaluate the representations by fine-tuning the whole network with few shot labels. Following the evaluation protocol in \cite{chen2020simple}, \cite{chen2020big}, we fine-tune all layers with only $1\%$ and $10\%$ labeled data. For fair comparison, we use the same splits of training data as in \cite{chen2020simple}, using SGD optimizer with momentum 0.9 to fine-tune all layers for 60 epochs. The initial learning rate is set to $10^{-4}$ for backbone and 10 for randomly initialized fc layer. During fine-tuning, only random cropping and horizontal flipping are applied for fair comparisons. Note that our method does not apply any special design like \cite{chen2020big}, which makes use of more MLP layers and has shown improved results when fine-tuning with few labels. As shown in Table \ref{tab:semi_imagenet}, our method achieves $62.2\%$ top-1 accuracy with only $1\%$ labels and $72.9\%$ top-1 accuracy with $10\%$ labels. In both two settings, our method consistently outperforms other semi-supervised and self-supervised methods, especially when $1\%$ labeled samples are available.

\begin{table*}[]
\renewcommand\arraystretch{1.2}
\setlength{\tabcolsep}{2.0mm}
\caption{Transfer learning accuracy (\%) on COCO detection and COCO instance segmentation (averaged over 5 trials).}
\vspace{0.05in}
\centering
\begin{tabular}{l|cccccc|cccccc}
\hline
\multirow{3}{*}{Method} & \multicolumn{6}{c|}{Mask R-CNN, R50-FPN, Detection}  &\multicolumn{6}{c}{Mask R-CNN, R50-FPN, Segmentation} \\ \cline{2-13}
                        % & \multicolumn{6}{c|}{$1\times$ schedule} & \multicolumn{6}{c}{$2\times$ schedule} \\ \cline{2-13}
& AP$^{\mathrm{bb}}$  & AP$^{\mathrm{bb}}_{\mathrm{50}}$ & AP$^{\mathrm{bb}}_{\mathrm{75}}$ & AP$_{\mathrm{S}}$  & AP$_{\mathrm{M}}$ &AP$_\mathrm{L}$ &AP$^{\mathrm{mk}}$  &AP$^{\mathrm{mk}}_{\mathrm{50}}$  &AP$^{\mathrm{mk}}_{\mathrm{75}}$  &AP$_\mathrm{S}$  &AP$_\mathrm{M}$ &AP$_\mathrm{L}$ \\ \hline

Supervised                  &38.9  &59.6  &42.0  &23.0  &42.9  &49.9    &35.4  &56.5  &38.1  &17.5  &38.2 &51.3     \\ \hline
MoCo v2 \cite{chen2020improved} &39.2  &59.9 &42.7 &23.8 &42.7 &50.0    &35.7  &56.8  &38.1  &17.8  &38.1  &50.5   \\
BYOL \cite{grill2020bootstrap} &39.9 &60.2 &43.2 &23.3 &43.2 &\textbf{52.8}   &- &-	&- &- &- &-\\
DenseCL\cite{wang2020dense}  &\textbf{40.3} &59.9 &\textbf{44.3} &- &- &-    &36.4 &57.0 &\textbf{39.2} &- &- &- \\ \hline

CsMl(w/o multi-level) &39.5 &60.1 &43.1 &24.1 &42.8 &50.9  &35.9 &56.9 &38.6 &18.3 &38.1 &\textbf{51.4} \\
CsMl(w/ multi-level) &\textbf{40.3}  &\textbf{61.1} &43.8 &\textbf{25.0} &\textbf{43.6} &50.8 &\textbf{36.6}  &\textbf{58.1}  &39.1  &\textbf{18.8}  &\textbf{39.0}  &51.2 \\ \hline
\end{tabular}
\label{tab: coco_detec}
\end{table*}

Following semi-supervised learning setting that the unlabeled samples are used for training via assigning pseudo labels, we further conduct a simple data mining procedure to access how the pre-trained models improve data mining. The data mining procedure is conducted as follows: 1) Using the fine-tuned model to infer on all unlabeled samples, and obtaining a confidence distribution for each image. 2) The information entropy is calculated to measure the confidence degree of each image, and we filter those samples with entropy higher than a threshold. 3) Convert the soft labels into one-hot pseudo label via only retaining the highest scored dimension, and train the model together with the ground truth labeled data.
We simply set the entropy threshold as 1 and the model is trained following a standard fully supervised learning procedure \cite{he2016deep}. Specifically, the model is trained for 90 epoch with initial learning rate set as $10^{-4}$ for backbone and $10^{-2}$ for fc layer, and they are both decayed by 0.1 after every 30 epochs. The results are shown in the last row of Table \ref{tab:semi_imagenet}, the performance can be further boosted via a simple data mining procedure, and we achieve $66.7\%$ top-1 accuracy with $1\%$ labeled samples, and $75.1\%$ top-1 accuracy with $10\%$ labeled samples. %Importantly, with only $10\%$ labels, we are approaching the fully supervised baseline ($76.5\%$) that makes use of all labeled samples.

\begin{table}[]
\caption{Transfer Learning results on PASCAL VOC detection (averaged over 5 trials).}
\vspace{0.05in}
\centering
\setlength{\tabcolsep}{5mm}
\begin{tabular}{lcccc}
\toprule
Methods & AP50 & AP75   \\
\midrule
Supervised                  & 81.4  & 58.8  \\
MoCo v2 \cite{chen2020improved}                    & 82.5  & 64.0  \\
CsMl                        & \textbf{82.7}   & \textbf{64.1}        \\
\bottomrule
\end{tabular}
\label{tab: voc_detec}
\end{table}

\begin{figure}
    \centering
    \includegraphics[width=0.75\linewidth]{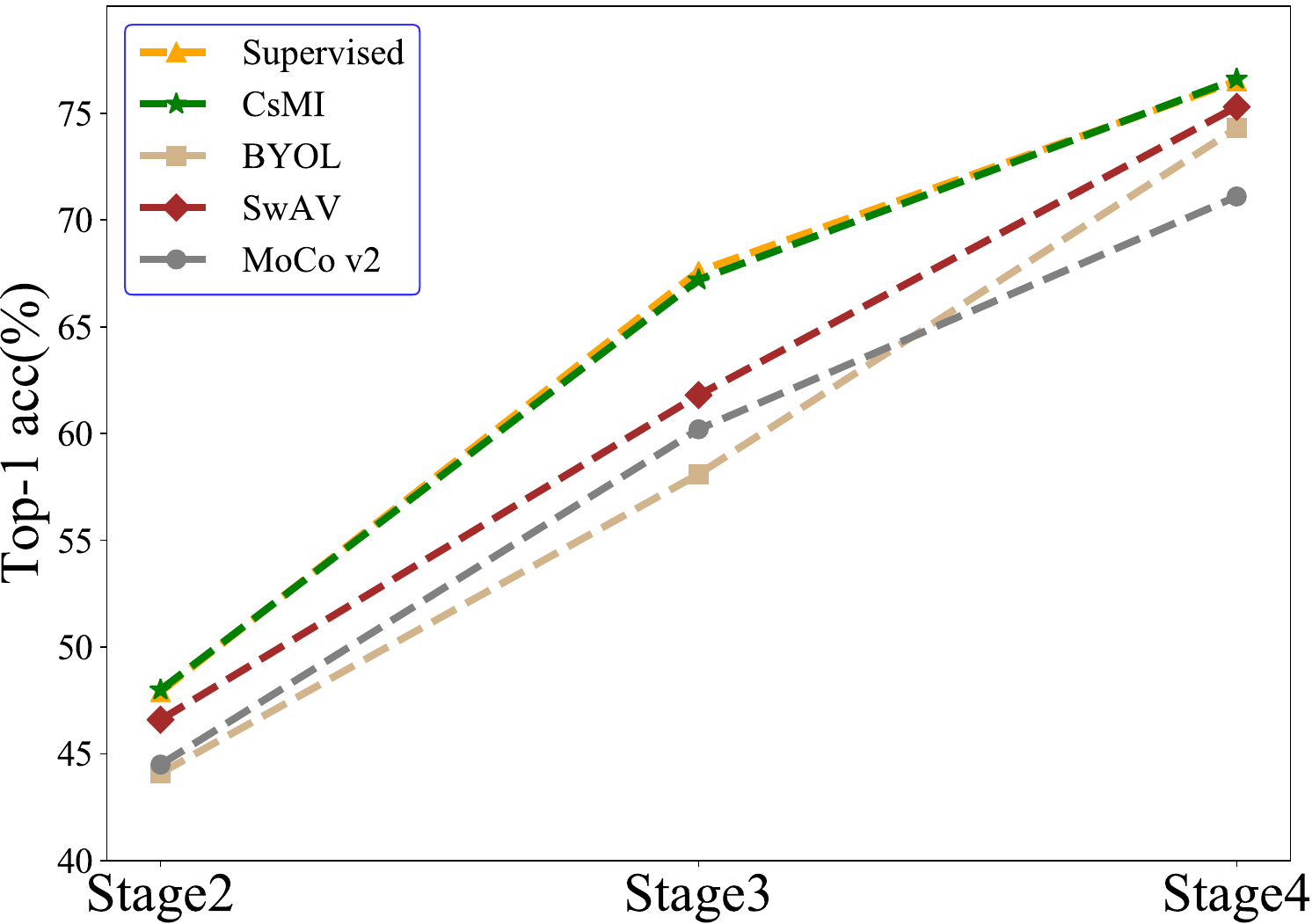}
    \vspace{0.05in}
    \caption{Top-1 linear classification accuracy of different methods at different stages using ResNet-50 as backbone.}
    \label{fig:inter_layer}
    \vspace{-0.2in}
\end{figure}

\subsection{Downstream Tasks}
%\vspace{-0.1in}
We also test the generalization of our unsupervised learned representations on more downstream tasks, including object detection and instance segmentation. All experiments follow MoCo \cite{he2020momentum} settings for fair comparisons.

\vspace{-0.1in}
\paragraph{PASCAL VOC}
Following evaluation protocol in \cite{he2020momentum}, we use Faster-RCNN \cite{ren2015faster} detector with R50-C4 backbone. All layers are fine-tuned end-to-end on the union set of VOC07+12 for $2\times$ schedule, and we evaluate the performance on VOC test07. As shown in Table \ref{tab: voc_detec}, CsMl achieves $82.7\%$ and $64.1\%$ mAP under AP50 and AP75 metric, which is slightly better than the results of MoCo v2.

\vspace{-0.15in}
\paragraph{MS COCO} We also evaluate the representation learned on a large scale COCO dataset. Following \cite{he2020momentum}, we use mask R-CNN \cite{he2017mask} detector with FPN, fine-tune all the layers end-to-end over the train2017 set, and evaluate the performance on val2017.  As shown in Table \ref{tab: coco_detec}, which compare the detection and segmentation results under default $1\times$ learning schedule. Our method consistently outperforms the supervised and MoCo v2 baseline.
% Specially, in $2 \times$ schedule, our method surpasses supervised pre-trained model by $1.6\%$ and $1.4\%$ under detection and segmentation task, respectively.
Notably, we achieve much higher performance on small and medium objects. This is mainly due to the hierarchical alignment strategy that increases the discrimination power of the intermediate layers. Note that our method is comparable with recently proposed DenseCL \cite{wang2020dense}  which is particularly designed for detection task. %\textcolor{red}{The results of $2 \times$ schedule are in appendix}

%Comparing with classification, although our proposed method
\begin{table}[]
\centering
\caption{Results of adding different modules.}

\vspace{0.05in}
\begin{tabular}{lc}
\toprule
Method                                  & Accuracy (\%)  \\
\midrule
MoCo v2                                 & 67.5         \\
$q_a$ + $q_p$                     & 70.4        \\
%$q_a$ + $q_p^1$ + $q_p^2$              & 68.6        \\
$q_a$ + $q_p$ + $\hat{q}$                   & 71.6        \\
$q_a$ + $q_p$ + $\hat{q}$ + multi-crop      & 74.6        \\
\bottomrule
\end{tabular}
\label{tab: diff_modules}
\end{table}
% \begin{figure}
%     \centering
%     \includegraphics[width=0.75\linewidth]{LaTeX/fig/diff_module_acc.pdf}
%     \caption{Results of adding different modules.}
%     \label{fig:diff_modules}
% \end{figure}

\subsection{Ablation Studies}
% \subsubsection{Impacts of hyper-parameter $\lambda$ on performance}
% The hyper-parameter $\lambda$ in Eq.\ref{eq: Loss1} and Eq.\ref{eq: Loss2} balances the strength between intra-sample pulling and inter-sample pulling. When $\lambda$ is set to $1$, only intra-sample pulling is performed. Correspondingly, when $\lambda$ is set to $0$, only inter-sample pulling is performed. Figure xxx shows the top-1 accuracy under various $\lambda$.
In this section, we conduct extensive ablation studies to better understand how each component affects the performance. Unless specified, \textbf{all results are compared with models trained for 200 epochs for efficiency}, and we report the top-1 accuracy under linear evaluation protocol.
\vspace{-0.1in}
\paragraph{Effects of Different Modules.} We first diagnose how each component affects the performance, as shown in Table \ref{tab: diff_modules}, simply add one positive sample $\mathbf{q}_p$ boost the baseline MoCo v2 by $2.9\%$, and introducing mixed samples for pulling further improve the performance by another $1.2\%$. Following \cite{caron2020unsupervised}, we also adopt multi-crop augmentations, and the performance can be further improved by $3\%$.% \textcolor{red}{We also compare the results of different positive sample selection strategy. As shown in Table \ref{tab: diff_modules}, the performance of using knn selected positive sample outperform using k-means selected positive sample by $0.9\%$. It is because that k-means aims to minimize the global clustering loss, and will bring higher noise if using k-means to select local neighborhood sample.}

\vspace{-0.15in}
% \paragraph{Number of Positive Samples.}
% \paragraph{$K$ in KNN Positive Sample Selection}
\paragraph{Effects of $k$ in KNN Positive Sample Selection}

% \begin{table}[]
% \caption{Top-1 accuracy under different neighborhood sample selecting strategy}
% \vspace{0.1in}
% \centering
% \renewcommand\arraystretch{1.1}
% \setlength{\tabcolsep}{3.0mm}
% \begin{tabular}{lll}
% \toprule
% Method      & w/ Mix($\mathbf{x_i}, \mathbf{x_j}$) &w/o Mix($\mathbf{x_i}, \mathbf{x_j}$) \\ \midrule
% %k-means      &          \\ \midrule
% Moco v2 \cite{chen2020improved} & - & 67.5 \\
% knn(K=0)    & 69.9       & 68.6  \\
% knn(K=1)    & 71.3       & 70.1  \\
% knn(K=5)    & 71.4       & 70.3  \\
% knn(K=10)   & 71.6       & 69.5  \\
% knn(K=100)  & 71.2       & 68.9  \\ \bottomrule
% \end{tabular}
% \label{tab: select_sample}
% \end{table}

% \begin{figure}
%     \centering
%     \includegraphics[width=0.75\linewidth]{LaTeX/fig/diff_module_acc.pdf}
%     \caption{Results of adding different modules.}
%     \label{fig:diff_modules}
% \end{figure}

% \begin{figure}
%     \centering
%     \includegraphics[scale=.26]{fig/diff_topk_acc.pdf}
%     %  \vspace{0.10in}
%     \caption{Top-1 accuracy with different $k$ in knn.}
%     \label{fig:diff_K}

% \end{figure}

%\paragraph{Effects of adding more positives.}
%\textcolor{red}{Refer to appendix}
% 把三个图直接放在一张pdf上，并标下标a,b,c, 然后总的title 直接 a,b,c 就可以了
\begin{figure*}
\centering
    % \subfigure[Results of adding different modules.]
    % {
    % \begin{minipage}{0.22\linewidth}
    % \centering
    %     \includegraphics[width=0.85\linewidth]{LaTeX/fig/diff_module_acc.pdf}
    %     % \caption{Results of adding different modules.}
    %     \label{fig:diff_modules}
    % \end{minipage}
    % }
    \subfigure[]%Top-1 accuracy with different $k$ in knn.
    {
    \begin{minipage}{0.32\linewidth}
    \centering
        \includegraphics[width=0.90\linewidth]{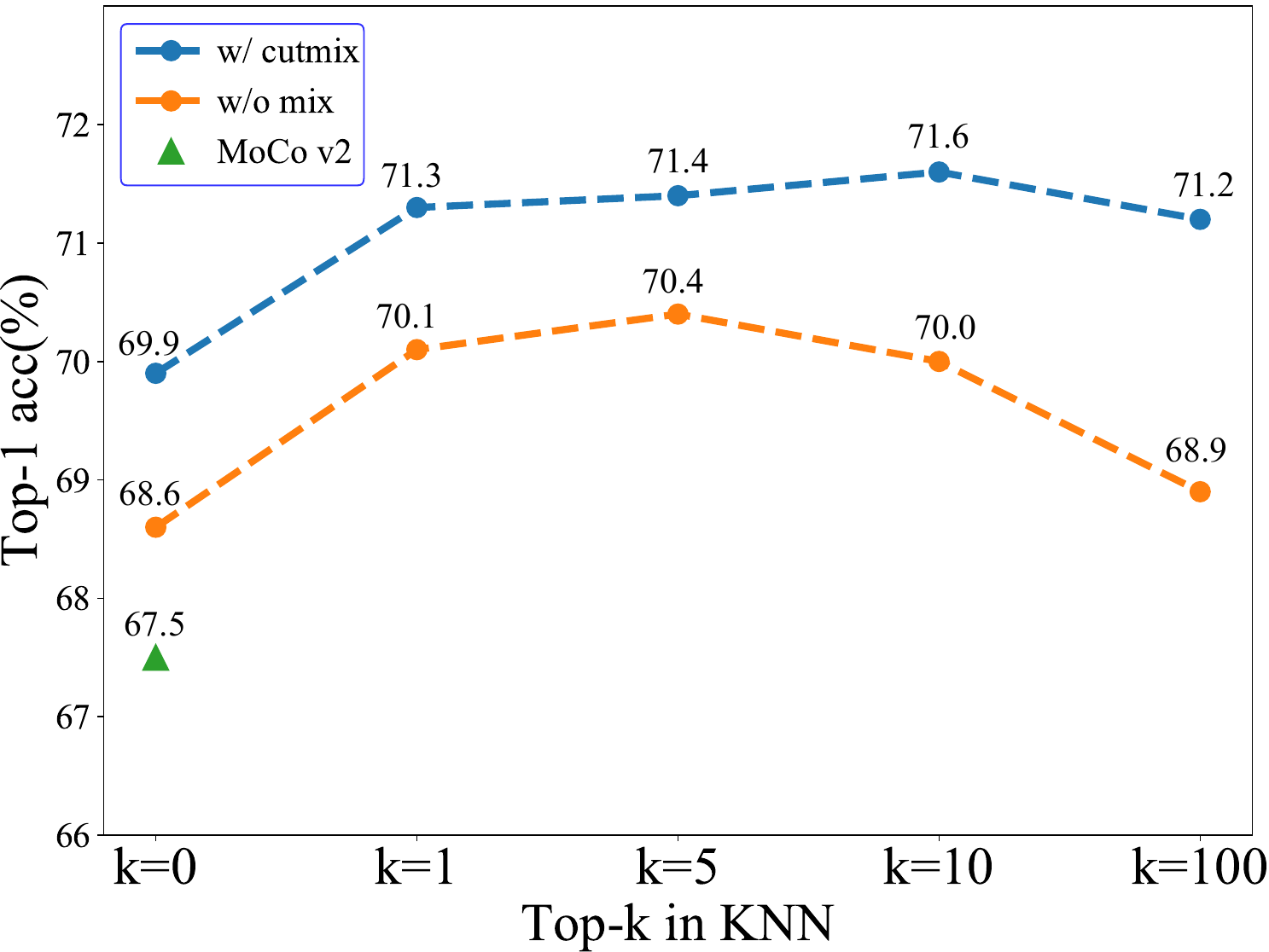}
        % \caption{Top-1 accuracy with different $k$ in knn.}
        \label{fig:diff_K}
        \vspace{0.07in}
    \end{minipage}
    }
    \subfigure[]%Top-1 accuracy and computational complexity comparisons for different models.
    {
    \begin{minipage}{0.32\linewidth}
    \centering
    \vspace{0.01in}
        \includegraphics[width=0.95\linewidth]{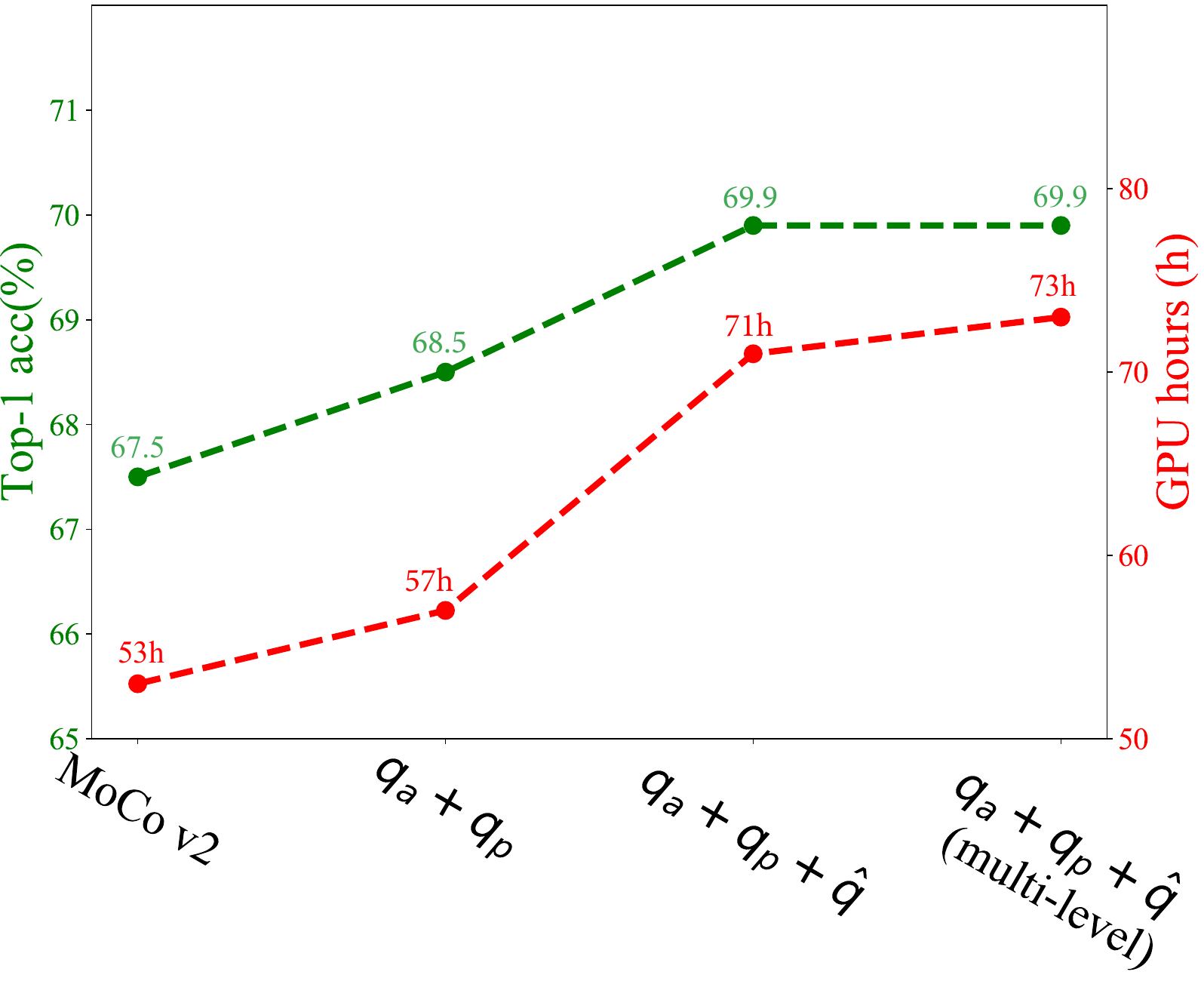}
        % \caption{Top-1 accuracy and computational complexity comparisons for different models.}
        \label{fig:gpu_hours}
    \vspace{-0.15in}
    \end{minipage}
    }
    \subfigure[]%Top-1 accuracy comparisons for different training epochs of CsMl and MoCo v2.
    {
    \begin{minipage}{0.32\linewidth}
    \centering
        \includegraphics[width=0.90\linewidth]{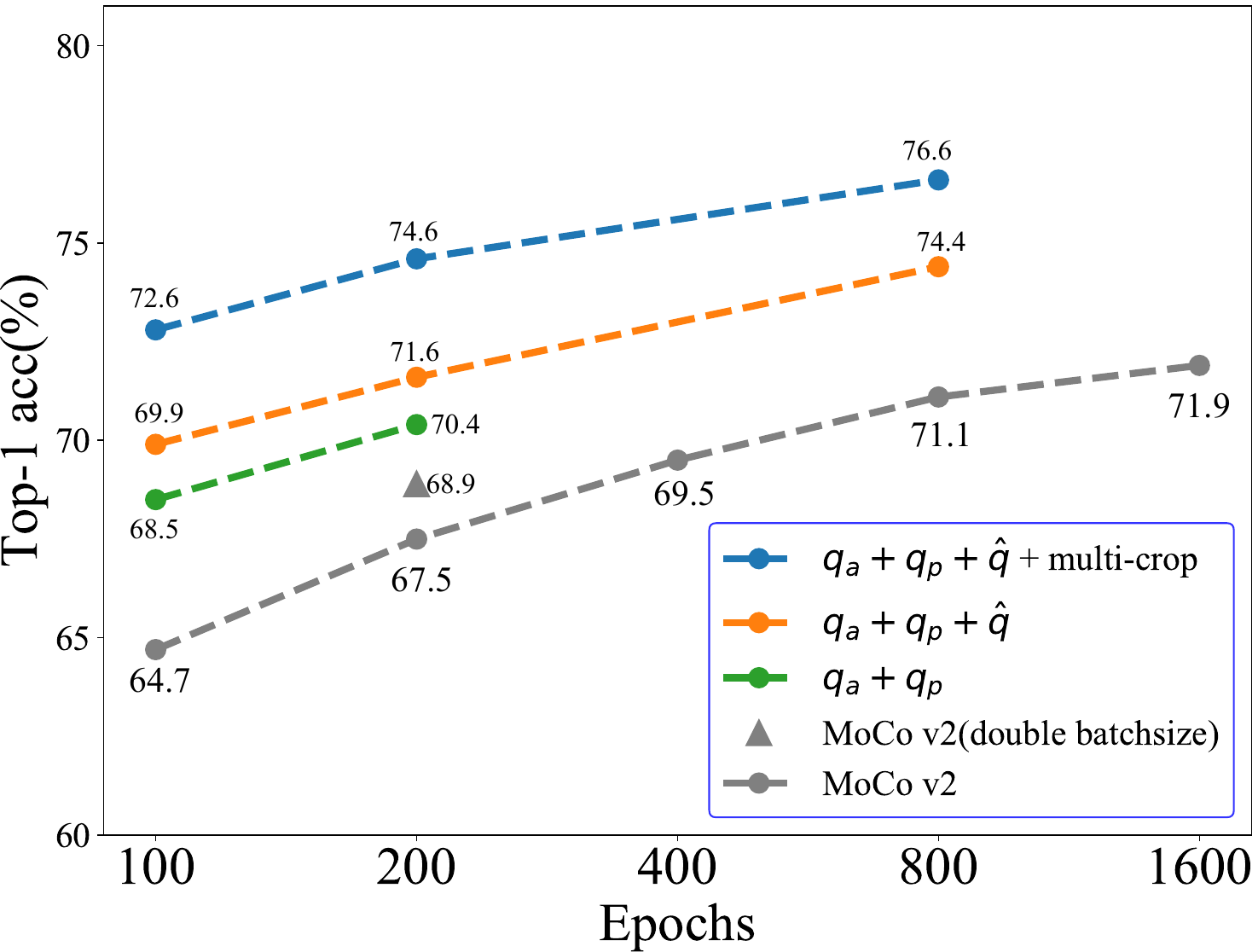}
        % \caption{Top-1 accuracy comparisons for different training epochs of CsMl and MoCo v2.}
        \label{fig:diff_epoch}
        \vspace{0.07in}
    \end{minipage}
    }
    \caption{(a) Top-1 accuracy with different $k$ in knn. (b) Top-1 accuracy and computational complexity comparisons for different models. (c) Top-1 accuracy comparisons for different training epochs of CsMl and MoCo v2.}

\end{figure*}

We then inspect the influence when selecting different number of samples as positive candidates. Fig. \ref{fig:diff_K} shows the top-1 accuracy with respect to different $k$ in knn. It can be shown that when no mixed sample is included, the performance is relatively sensitive to $k$, and the results are relatively robust for a range of $k$ ($k=1,5,10,100$) when mixed samples are introduced. The reason is that mixed samples act as a strong smoothing regularization, which alleviates the noise introduced by knn selection. We also consider a special case, \emph{i.e.,} always select the same sample, \emph{i.e.,}, $\mathbf{x}_p=\mathbf{x}_a$, noted as $k=0$. In this setting, we always select the correct positive sample for pulling, but sacrifice the diversity of using cross-samples. The performance gain is limited ($67.5\% \rightarrow 68.6\%$), which validate the effectiveness of selecting positives via cross samples.

\vspace{-0.15in}
\paragraph{Effects of Pulling Multi-level Views.}
\begin{table}[]
\renewcommand\arraystretch{1.1}
\setlength{\tabcolsep}{1.8mm}
\caption{Classification accuracy with features of different stages and fine-tuning results with $1\% $ labels.}
\vspace{0.05in}
\begin{tabular}{lcccc}
\toprule
\multicolumn{1}{c}{} & Stage2 & Stage3 & Stage4 & 1\% label \\
\midrule
MoCo v2 \cite{chen2020improved}             & 44.5   & 60.2   & 71.1   & 52.4      \\
Pull stage4          & 45.1   & 60.7   & \textbf{71.6}   & 53.4      \\
Pull stage3\&4       & 45.6   & \textbf{64.0}   & 71.5   & 54.5      \\
Pull stage2\&3\&4    & \textbf{46.0}   & \textbf{64.0}   & \textbf{71.6}   & \textbf{54.8}      \\
\bottomrule
\end{tabular}
\label{tab: mulhead}
\end{table}

We analyze the performance of the intermediate layers and validate how our proposed hierarchical alignment strategy benefits the representation. Except for the last layers of stage 4, we also explicitly pull similar samples in earlier layers such as stage 2 and stage 3. Table \ref{tab: mulhead} shows the linear classification accuracy of each stage, it can be shown that pulling similar samples in the shallow layers consistently increases its discrimination power, especially for stage 3, \emph{e.g.,} the accuracy increased by $3.3\%$, from $60.7\%$ to $64.0\%$. The performance gain is relatively small in stage 2, partially because the representation in this stage is too low-level, and is hard for global representation. We also note that the performance of the last layers is not significantly affected by penalizing the shallow layer. However, We find that better separability in the shallow layers is beneficial for fine-tuning $1\%$ labels, the accuracy increased by $1.4\%$ when introducing shallow pulling. The advantages of increased representation of shallow layers can be also validated by the downstream detection and segmentation tasks, as shown in Table \ref{tab: coco_detec}. We also compare the linear classification accuracy of different methods at intermediate layers, as shown in Fig. \ref{fig:inter_layer} %(这个fig有点远，你调整一下位置)
% and Table \ref{tab: coco_seg}.

%comparing with no although the last layer's top-1 accuracy is almost same no matter explicate pulling on shallow layer is performed or not, shallow layer's top-1 accuracy is improved obviously. It means that the quality of shallow layers' representation get better. In this way, downstream tasks will be benefit from better shallow representations, as shown in the last column in Table \ref{tab: mulhead}, the $1\%$ fine-tune results increase by $1.4\%$ compared with only pulling samples in the last layer.

%To verify the efficiency of shallow layer's feature, we perform experiments on COCO detection downstream task with FPN backbone. As shown in Table xxx, better shallow feature also improve the performance of detection.
\subsection{Analysis and Discussions}

\paragraph{Analysis of epoch\& batch size} In our implementation, the batch size is measured based on the number of anchors, and the actual samples are doubled during each forward propagation when we select one positive sample for pulling. In order to get rid of the effects of batch size, we compare our methods with MoCo v2 that trained for double epochs. The results are 68.5\% and 70.4\% for 100 and 200 epochs, which is better than MoCo v2 with accuracies of 67.5\% and 69.5\% for 200 and 400 epochs, respectively. Note that, Moco v2 achieves marginal improvement when extending the training epochs from 800 to 1600 ($71.1\% \rightarrow 71.9\%$), while the proposed method achieves much better (74.4\% for 800 epochs) performance. In addition, we double the batch size of MoCo v2 and train it for 200 epochs, the actual batch size of MoCo is same as our method in such setting. As shown in Fig. \ref{fig:diff_epoch}, the performance gain of enlarging batch size is $1.4\%$ (($67.5\% \rightarrow 68.9\%$)), which is lower than that of CsMl under same epochs (70.4\%). The results verify that the improvement of CsMl is not simply caused by extending the training epochs.

%In our methods, there are multiple queries during each forward-backward propagation, \emph{i.e.,} $\mathbf{x_a}$, $\mathbf{x_p}$, and mixed sample $\hat{x}$, which is unfair to compare with other methods that only have one query. In order to inspect whether the performance gain is mainly from enlarging the batch size of the query, we compare with MoCo v2 results with more training epochs. As shown in Fig. \ref{fig:diff_epoch}, \textcolor{red}{When mixed image is not involved, our method with 100 and 200 training epochs outperform moco v2 of 200 and 400 epochs. After use mixed images, our method with 200 epochs pretrian even outperform moco v2 with 800 epochs.}
% With 200 training epochs, our method outperforms moco v2 with 600 epochs by $1.1\%$ and even better than moco v2 with 800 epochs.
%This demonstrates the effectiveness of pulling semantically similar samples and the performance improvement is not simply from increasing the batch size of positive samples.
\vspace{-0.15in}
\paragraph{Computational Complexity} We compare the computational complexity with MoCo v2 under 200 epochs pretraining and different variants of CsMl under 100 epochs pretraining with 8 v100 GPUs. As shown in Figure \ref{fig:gpu_hours}, when mixed samples are not involved, the cost time of MoCo v2 and CsMl is almost the same: 53h v.s. 57h. The increased 4 hours is mainly from the cost of nearest sample selection. When the mixed sample is included in the query, the computation cost increased by roughly $25\%$. Finally, we find that the computation cost of multi-level is marginal because the additional bottleneck's computation is very small.

\section{Conclusion}
This paper proposes a hierarchical training strategy that pulls semantically similar images  for contrastive learning. The main contributions are two folds, first, in order to select similar samples without labels, we deliberately design a sample selection strategy relying on data mixing, which generates new samples that current model does not perform well. The highlight is that samples are only mixed from those similar samples and does not destroy the local similarity structure. Second, we extend the semantic alignment to intermediate hidden layers and enforces the feature representation to be discriminative throughout the network. In this way, the network can be optimized in a more robust way and we find it is beneficial for general representation. We conduct extensive experiments on widely used self-supervised benchmarks, and consistently outperforms previous self-supervised learning methods.
{\small
\bibliographystyle{ieee_fullname}
\bibliography{egbib}
}

\input{appendix}

\end{document}

%% file: math_commands.tex
\usepackage{amsmath,amsfonts,bm}

\def\eqref#1{equation~\ref{#1}}
\def\1{\bm{1}}

\DeclareMathAlphabet{\mathsfit}{\encodingdefault}{\sfdefault}{m}{sl}
\SetMathAlphabet{\mathsfit}{bold}{\encodingdefault}{\sfdefault}{bx}{n}

%% file: appendix.tex
\clearpage
\appendix
\section{Appendix}

\renewcommand\thefigure{\thesection.\arabic{figure}}
\renewcommand\thetable{\thesection.\arabic{table}}
\setcounter{figure}{0} 
\setcounter{table}{0} 

\begin{table}[]
\centering
\caption{Top-1 accuracy comparisons based on BYOL}

\vspace{0.01in}
\begin{tabular}{lc}
\toprule
Method  & Accuracy(\%)  \\
\midrule
BYOL 300 epochs & 72.5        \\
BYOL 1000 epochs & 74.3        \\
CsMl 300 epochs & 75.3 \\

\bottomrule
\end{tabular}
\label{tab: HSA_byol}
\end{table}

\begin{table}[]
\centering
\caption{Top-1 accuracy comparisons with more positive pairs}

\vspace{0.01in}
\begin{tabular}{lc}
\toprule
Method  & Accuracy(\%)  \\
\midrule
$q_a + q_p$ & 70.4        \\
$q_a + q_{p1} + q_{p2}$ & 70.2  \\
$q_a + q_p + \hat{q}$ & 71.6 \\

\bottomrule
\end{tabular}
\label{tab: more_postive}
\end{table}

\subsection{{Results on BYOL}}
{Our method is applicable to other contrastive based methods to further improve the performance. As shown in \ref{tab: HSA_byol}, we applied CsMl to BYOL [10], which does not need negative samples during contrastive learning. Following the hyper-parameters used in [10], we pretrain the model with CsMl for 300 epochs. CsMl achieves an accuracy of $75.3\%$ linear classification accuracy, which is even better than BYOL under 1000 epochs pre-training. }

\subsection{{Results of more positive pairs}}
{Based on Eq. (2), CsMl can also be extended to support pulling multiple positive samples during each forward propagation. 
As shown in Table \ref{tab: more_postive}, using 2 positive pairs achieves similar result as using 1 positive pairs. As comparison, introducing mixed samples as positive pairs is more effective. Considering  that adding more positive samples would inevitably increasing the computational complexity, we simply choose one positive sample for each anchor. 
}

\subsection{MixUp v.s. CutMix augmentations}
\begin{table}[]
\centering
\caption{Comparisons of Mixup with CutMix augmentations.}

\vspace{0.01in}
\begin{tabular}{lc}
\toprule
Method                                            & Accuracy(\%)  \\
\midrule
MoCo v2                                             & 67.5         \\
$q_a$ + $q_p$                                       & 70.4        \\
$q_a$ + $q_p$ + $\mathbf{MixUp}(q_a, q_p)$          & 70.1        \\
$q_a$ + $q_p$ + $\mathbf{CutMix}(q_a, q_p)$         & 71.6        \\

\bottomrule
\end{tabular}
\label{tab: cutmix_mixup}
\end{table}

Besides CutMix, We also consider another cross-sample augmentation strategy Mixup [ref] to expand the neighborhood of an anchor. As shown in Table \ref{tab: cutmix_mixup}, Mixup is worse than Cutmix, and even slightly worse than baseline method that do not involve any mixed sample in query, \emph{i.e.} $q_a + q_p$ setting, the reason may be that Mixup augmentations destroy the naturality of the pixel distribution. 

\subsection{Visualization of Feature Representation} 
\begin{figure}[t!]
  \begin{center}
        \includegraphics[width=0.98\linewidth]{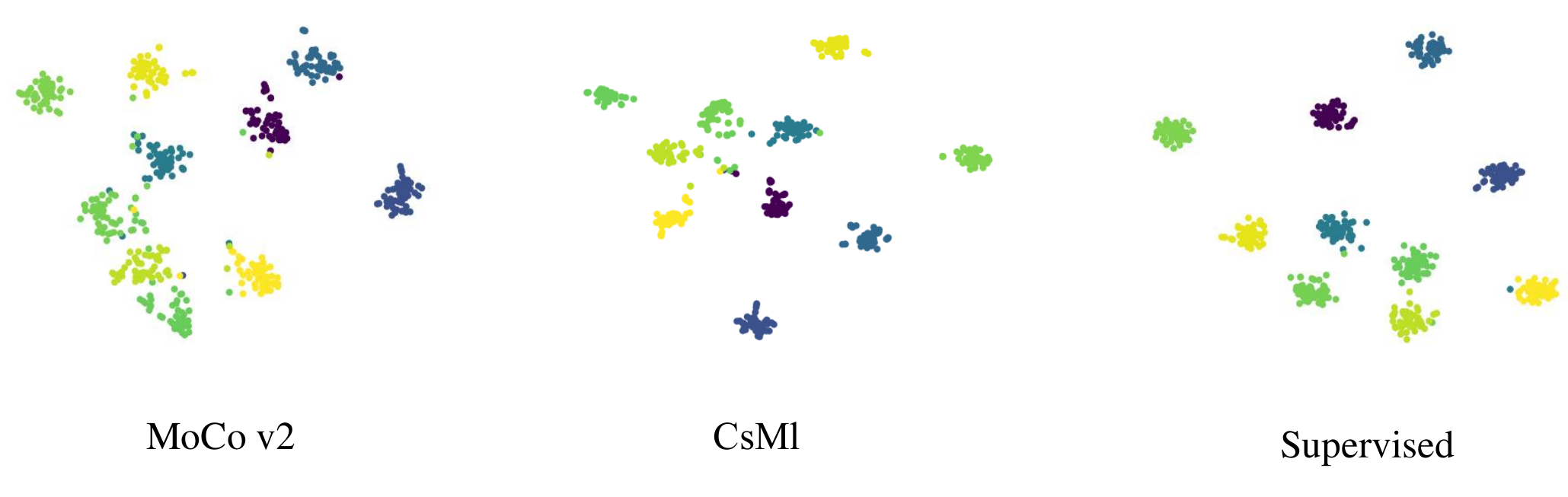}
  \end{center}
  \vspace{-0.3cm}
     \caption{\textit{t-sne} visualization of representation learned by MoCo v2, CsMl, and fully supervised learning.}
  \label{t-sne}
\end{figure}

We visualize the last embedding feature to better understand the semantic alignment properties of the proposed method. Specifically, we randomly choose 10 classes from the validation set and provide the \textit{t-sne} visualization of feature representation, supervised training and MoCo v2. As shown in Fig.~\ref{t-sne}, the same color denotes features with the same label. It can be shown that CsMl presents higher alignment property comparing with MoCo, and the fully supervised learned representation reveals the highest alignment due to the available of image labels.
% \begin{figure}
%     \centering
%     \includegraphics[width=0.9\linewidth]{fig/diff_epoch.pdf}
%     %  \vspace{-0.1in}
%     \caption{Top-1 accuracy comparisons with different epochs of MoCo v2 and CsMl.}
%     \label{fig:diff_epoch}

% \end{figure}
% \begin{figure}
%     \centering
%     \includegraphics[width=0.75\linewidth]{fig/diff_epoch_acc.pdf}
%     \caption{Top-1 accuracy comparisons for different training epochs of CsMl and MoCo v2.}
%     \label{fig:diff_epoch}
% \end{figure}

% \subsection{The results of more training epochs}
% \begin{figure}
%     \centering
%     \includegraphics[width=0.75\linewidth]{fig/diff_epoch_acc.pdf}
%     \caption{Top-1 accuracy comparisons for different training epochs of HSA and MoCo v2.}
%     \label{fig:diff_epochs}
% \end{figure}

% To better understand the performance of our proposed HSA method, we train HSA and MoCo v2 for longer epochs to diagnose how the representation evolves for more training epochs. As shown in Fig. \ref{fig:diff_epochs}, when training for longer epochs, the improvement of HSA is trivial, \emph{i.e.,} only $0.1\%$ improvement when we extend the training epochs from 800 to 1200, the performance tends to converge for HSA. While for MoCo, the performance improves by $0.8\%$ when extending from 800 epochs to 1600 epochs, the convergence rate is rather slow due the pretext task in MoCo that only regards a single image as well as its augmentations as positive samples. 
% %In addition, we find that HSA will converge at around $76.4\%$. When we train HSA from 800 epochs to 1200 epochs, only $0.1\%$ improvement is brought. Note that the results of HSA in Figure \ref{fig:diff_epochs} is trained with multi-crop data augmentation, which is slightly different from Figure 3 in Section 4.4.

\subsection{Performance comparisons of CsMl and supervised model}
% \begin{table}[]
% \caption{Distribution on some categories of HSA and fully supervised model. The categories with bold font is ground truth, and below is the highest-scoring error category}
% \centering

% \begin{tabular}{llll}
% \toprule
% Label & Semantic Label    & Supervised & HSA  \\
% \midrule
% \textbf{543}   & \textbf{dumbbell}          & 60.0       & 84.0 \\
% 422   & barbell           & 24.0       & 4.0  \\
% \hline
% \textbf{587}   & \textbf{hammer}            & 50.0       & 58.0 \\
% 784   & screwdriver       & 8.0        & 2.0  \\
% \hline
% \textbf{250}   & \textbf{Siberian husky}    & 66.0       & 58.0 \\
% 248   & Eskimo dog, husky & 22.0       & 30.0 \\
% \hline
% \textbf{35}    & \textbf{mud turtle}        & 70.0       & 54.0 \\
% 36    & terrapin          & 8.0          & 28.0 \\
% \bottomrule
% \end{tabular}
% \label{tab: dist_all_class}
% \end{table}

\begin{figure*}
    \centering
    
    \includegraphics[width=0.98\linewidth,height=9cm]{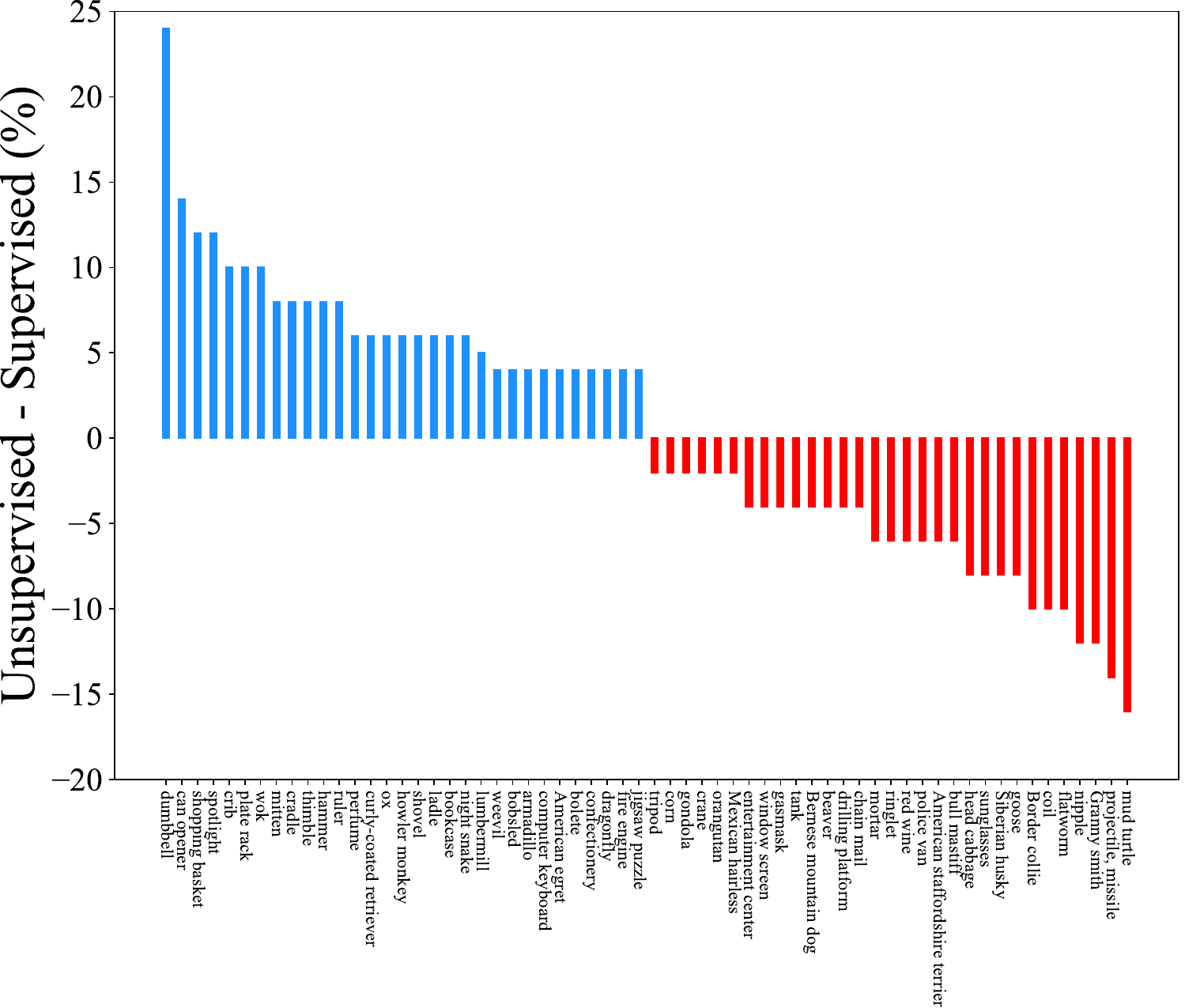}
    \caption{Top-1 accuracy comparisons on different categories of CsMl and fully supervised model. Here we compare most successful categories of each method.}
    \label{fig:diff_sup_unsup}
\end{figure*}

\begin{figure*}
    \centering
    
    \includegraphics[width=1.0\linewidth]{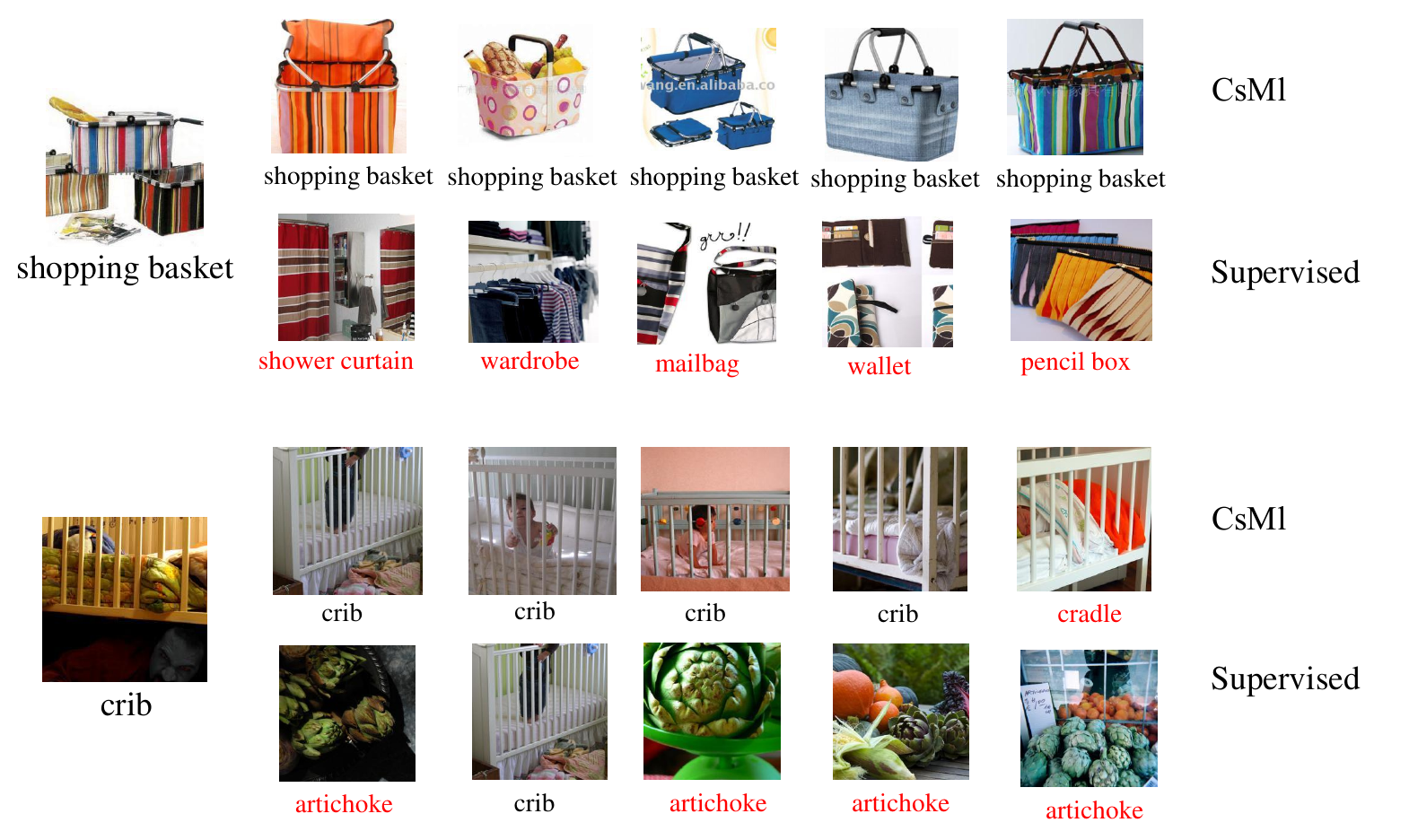}
    \caption{An illustration of the selected knn samples using CsMl and the supervised model.}
    \label{fig:diff_sup_unsup}
\end{figure*}

Here we diagnose the performance difference of CsMl and fully supervised baseline to uncover the advantages of each model. The performance evaluation is under the linear classification protocol and we report per-category accuracy. As shown in Fig. \ref{fig:diff_sup_unsup}, we illustrate the most successful categories of each model, and find that the advantage of supervised model lies in discriminating fine-grained subcategories, \emph{e.g.}, on many sub-classes of dogs (Border collie, Siberian husky, bull mastiff \emph{etc.}). It is intuitive since feature representation among fine-grained sub-categories is very similar, and it is hard to discriminate them without ground truth labels. While for unsupervised model CsMl, the most successful categories roughly around categories that require contour information for discrimination, \emph{e.g. shopping basket and plate rack}, while supervised model usually focuses on discriminative details such as texture. Following the evaluation in Table 2 in the original paper, we also show some example images that use knn for similar samples selection, \emph{i.e.,} given an image in the validation set, and find its top-k nearest neighbors in the training set. The most successful categories of CsMl results from capturing the global contour information.

%\section{Visualization of KNN classification}
%We present nearest neighbors in the training set for some validation samples in Figure \ref{}. We compare the visualization results with fully supervised model. 